\documentclass[journal]{IEEEtai}

\usepackage{color,array}
\usepackage{amsmath}
\usepackage{float}
\usepackage{amsfonts}
\usepackage{amssymb}
\usepackage[font=small,labelfont=bf,labelsep=colon,justification=centering]{caption}

\usepackage{graphicx}
\usepackage{float}
\usepackage{multirow}
\usepackage{threeparttable}
\usepackage{xcolor}

\begin{document}

\title{Insider Threat Detection Using GCN and Bi-LSTM with Explicit and Implicit Graph Representations} 

\author{Rahul Yumlembam, Biju Issac, Seibu Mary Jacob, Longzhi Yang and Deepa Krishnan
\thanks{Rahul Yumlembam, Biju Issac and Longzhi Yang are with the School of Computer Science, Northumbria University, Newcastle, UK (email: r.yumlembam@northumbria.ac.uk, bissac@ieee.org, longzhi.yang@northumbria.ac.uk). Corresponding author: Biju Issac}
\thanks{Seibu Mary Jacob is with the School of Computing, Engineering \& Digital Technologies, Teesside University, Middlesbrough, UK (email: s.jacob@tees.ac.uk).}
\thanks{Deepa Krishnan is with the Mukesh Patel School of Technology, Management, and Engineering, NMIMS University, India (email: Deepa.Krishnan@nmims.edu).}
}


\markboth{Journal of IEEE Transactions on Artificial Intelligence, Vol. 00, No. 0, Month 2020}
{First A. Author \MakeLowercase{\textit{et al.}}: Bare Demo of IEEEtai.cls for IEEE Journals of IEEE Transactions on Artificial Intelligence}

\maketitle

\begin{abstract}
Insider threat detection (ITD) remains a significant challenge in cybersecurity due to the concealed nature of malicious activities by trusted insiders. This paper introduces a novel post-hoc ITD framework that enhances detection capabilities by integrating explicit and implicit graph structures with a temporal component to analyse user behaviour effectively. We construct an explicit graph using predefined rules that capture user activities within an organisation's network, providing insights into explicit relationships between actions. To address potential noise and sub-optimality in the explicit graph, we complement it with an implicit graph derived from feature similarities using the Gumbel-softmax trick, which refines the structure by leveraging underlying patterns. Both graphs are processed through separate Graph Convolutional Networks (GCNs) to produce node embeddings, which are then concatenated and refined using an attention mechanism to emphasise critical features for threat detection. These refined embeddings are subsequently fed into a bidirectional Long Short-Term Memory (Bi-LSTM) network to capture the temporal dynamics of user behaviour. The model flags activities as anomalous if their probability scores fall below a predefined threshold. Extensive evaluations on two CERT datasets, r5.2 and r6.2, demonstrate that our framework significantly outperforms state-of-the-art methods. On the r5.2 dataset, our model achieves an Area Under the Curve (AUC) of 98.62, a perfect Detection Rate (DR) of 100\%, and a low False Positive Rate (FPR) of 0.05. For the more challenging r6.2 dataset, the model achieves an AUC of 88.48, a DR of 80.15\%, and an FPR of 0.15. These results illustrate that combining explicit and implicit graph representations, along with advanced sequential modelling, leads to a robust ITD solution capable of effectively distinguishing between normal and abnormal activities, even in complex scenarios.

\end{abstract}



\section{Impact Statement}

\IEEEPARstart{T}{he} 2024 Insider Threat Report by Cybersecurity Insiders \cite{Nadeau} reveals that 83\% of organizations experienced at least one insider attack in the past year. Even more striking is the surge in organizations reporting 11-20 insider attacks, which increased fivefold compared to 2023 — rising from just 4\% to 21\% within the last 12 months. Insider threats are one of the most critical challenges in modern cybersecurity. Trusted individuals with legitimate access to sensitive systems can execute malicious activities that are notoriously difficult to detect using conventional methods. Current approaches often rely on either explicit graphs, which capture observable relationships, or implicit graphs, which identify hidden patterns. However, each alone has limitations, leading to suboptimal detection performance. 

The framework introduced in this paper overcomes these limitations by combining explicit and implicit graph representations with temporal modeling via a Bi-LSTM network. This hybrid approach captures both observable and hidden relational patterns while analyzing the sequential dynamics of user behaviour. Tested on the CERT insider threat datasets, the framework achieved groundbreaking results. In the r5.2 data set, it reached an AUC of 98.62, a detection rate of 100\%, and a false positive rate of 0.05, surpassing state-of-the-art methods such as LAN and DeepLog. On the more challenging r6.2 dataset, it demonstrated robust performance with an AUC of 88.48 and a detection rate of 80 15\%.

This research has the potential to significantly enhance insider threat detection systems in environments such as critical infrastructure, financial services, and government organizations, where timely and accurate detection is essential to safeguard sensitive data and assets. By improving detection accuracy and reducing false positives, this work lays a strong foundation for developing more reliable and efficient solutions to address insider threats.

\section{Introduction}

The increasing threat of malicious insider activities within organisations is a critical concern. These threats are challenging to detect due to insiders' legitimate network access. Organisations routinely collect network logs encapsulating various information - login/logout times, opened files, removable device connection/disconnection, web browsing history, and email communication. Analysing these logs can reveal consistent patterns in a user's daily behaviour, considering the assumption that a user's activities are mainly consistent. Numerous studies have been conducted on insider threat detection, which can generally be categorised into three main types. The first category focuses on identifying various patterns of insider threats and conducting anomaly detection. These methods aim to establish baseline behaviours for users to differentiate between normal users and potential insider threats, employing approaches such as machine learning and deep learning. The second category of methods emphasises transforming user behaviours into sequence data, capturing the temporal relationships among log entries. Techniques like recurrent neural networks (RNN) and long short-term memory  (LSTM) are frequently used to capture temporal aspects of activities.
The third group constructs graphs that model the relationships between users or activities, capturing their underlying relational structure.

In this work, we focus on combining graph-based and temporal methods due to their ability to capture complex and subtle patterns in relational data, along with the temporal aspects of the data. Most previous work has approached the problem by either constructing implicit graphs or explicit graphs. In modelling relationships among activities, recent studies have approached implicit graph structure learning as a process of learning similarity metrics within the node embedding space. This approach assumes that node attributes inherently contain valuable information for deducing the implicit topological structure of the graph \cite{cai2024lan}. In explicit graph learning, recent work constructs the graph based on predefined rules that consider different relationships between activities \cite{GCN-Jiang}, \cite{zheng2022insider}, \cite{Adasage}, \cite{Log2vec}. This method captures direct, observable interactions, providing a clear and structured representation of user activities and interactions.

Relying solely on either an implicit graph or an explicit graph poses significant challenges. Using only an implicit graph assumes that its optimized structure represents a "variation" or substructure tailored for the downstream task. Graph structure learning aims to capture this optimized graph through a similarity metric, but it might need to include the valuable contextual information embedded in explicit relationships. On the other hand, an explicit graph captures direct, observable interactions but may overlook hidden patterns and subtle relationships crucial for detecting anomalies. Explicit graphs often contain noise and may be incomplete, thus providing an inadequate representation of the data. In contrast, the implicit graph, derived from feature similarities, is optimized for downstream prediction tasks and can identify refined, optimized relationships that the explicit graph might overlook.

Therefore, this work combines both explicit and implicit graphs to leverage the strengths of each, ensuring a comprehensive representation of the data. By integrating explicit graphs, we capture direct, observable interactions, while implicit graphs reveal hidden patterns and subtle relationships optimized for downstream tasks. Additionally, we recognize that logical relationships between activities are insufficient on their own because activities occur sequentially; thus, considering the temporal aspect is also crucial. To capture the temporal aspect, we implement a Bi-LSTM on the node embeddings generated after applying an attention mechanism to the node embeddings generated from both the explicit and implicit graphs using a GCN. This approach ensures that we effectively capture both the relational and temporal dynamics of user activities, enhancing our ability to detect anomalies and insider threats accurately. In summary, the contribution of this work is as follows:

\begin{itemize}
    \item  We propose a novel approach that combines explicit and implicit graph structures for insider threat detection. This integration leverages the strengths of both graph types, capturing both direct, observable interactions and refined, optimised relationships derived from feature similarities.
    \item Our framework employs an attention mechanism to refine the node embeddings generated from both explicit and implicit graphs. This attention layer helps to emphasise the most relevant features for threat detection, enhancing the model's overall performance.
    \item  We incorporate a bidirectional Long Short-Term Memory (Bi-LSTM) network to capture the temporal dynamics of user activities. This addition allows our model to analyse sequential patterns in the data, improving the detection of anomalies.
\end{itemize}

\section{Related works}
The study of insider threat detection extensively uses machine learning, considering various scenarios and evaluation methods. A typical training method involves feature extraction from logs, with features being either aggregations over time or temporal features. For instance, \cite{granularity-Le} trains multiple supervised machine learning algorithms using feature aggregation at session, day, week, and month levels. Similarly,  \cite{DeepMIT}, \cite{sun2021insider}, \cite{unsupervised-Le} aggregates features across weeks and months to train various unsupervised machine learning algorithms, flagging anomalies using an ensemble of outputs. Studies like DeepMIT \cite{DeepMIT} and  \cite{he2021insider} train sets of Recurrent Neural Networks (RNN) with unique hidden states for each user using temporal features.
Conversely, \cite{Lstm-auto-sharma} employ an LSTM Autoencoder for anomaly detection, using session-based feature aggregation over a fixed time window. However, the aggregated features in these studies \cite{granularity-Le}, \cite{unsupervised-Le} result in a loss of granularity, as forming a feature requires combining multiple logs. Furthermore, work that employs aggregated and temporal features frequently overlooks the relationship between logs and events.

To address this, researchers have proposed various graph-based approaches that capture the causal relationships between logs to identify anomalies. Graph-based algorithms such as Graph Convolutional Networks (GCN) have been applied to both homogeneous \cite{GCN-Jiang} and heterogeneous explicit graphs \cite{zheng2022insider}, constructed based on relationships between various entities. The study Adasage \cite{Adasage} proposes an edge-centric detection approach, constructing graph edges representing interactions between entities. Log2vec \cite{Log2vec} suggests constructing a heterogeneous graph based on ten different rules, followed by node embedding generation using random walk algorithms and Word2Vec models, which are then clustered to identify malicious users. Work in LAN \cite{cai2024lan} constructs implicit graphs by retrieving the K nearest activities from the activity vector pool and constructs an activity graph used to predict masked activity probabilities, showing promising results.

In addition to graph-based methods, researchers have explored time-series classification for insider threat detection. For example, \cite{Chattopadhyay2018} construct time-series feature vectors from daily user activity logs and address class imbalance through cost-sensitive undersampling of nonmalicious instances. A two-layered deep autoencoder is then employed and compared against random forest and multilayer perceptron, with experiments on the large-scale CMU Insider Threat dataset showing that the autoencoder and random forest achieve high precision, recall, and F-score. However, while the multilayer perceptron provides higher recall, it suffers from lower precision and overall F-score, indicating limitations in effectively balancing detection performance.

More recently, \cite{xiao2024unveiling} introduced a comprehensive framework that integrates both statistical and sequential information for insider threat detection. Their approach, termed CATE, employs convolutional attention to capture statistical features and a transformer encoder to learn sequential dependencies in user activity logs. Evaluation on the CERT dataset demonstrates that this dual-module design enhances robustness and accuracy compared to prior methods that considered either dimension in isolation. Building on these advancements, \cite{BiLSTM-Manoharan}  recurrent neural network architectures have also been employed to model sequential dependencies in insider activities. A notable example is the BiLSTM-based approach, which captures bidirectional dependencies in user behavior sequences and has been shown to outperform conventional RNN and LSTM architectures on the CERT r4.2 dataset by effectively combining manual features, sequential features, and ground truth labels. However, this approach is constrained by its reliance on historical daily records, which may limit adaptability to evolving behavioral dynamics.

In line with these approaches, Wayne et al. \cite{hong2023graph} introduced an integrated feature engineering solution on the CERT 4.2 dataset, combining manually-selected and automatically-extracted features. They employed an LSTM auto-encoder to learn latent representations from sequential activities, and further enhanced detection performance with a residual hybrid network (ResHybnet) that integrates GNN and CNN layers with an organizational graph. Their model achieved an improvement of 0.56\% in F1 score through the LSTM auto-encoder and an additional 1.97\% performance boost with ResHybnet over existing baselines. However, their work is limited by its dependence on organizational graphs constructed from user-day nodes, which may not generalize well to settings with sparse relational structures or unseen activity patterns. In \cite{Manoharan2023Bilateral} the authors have introduced a bilateral insider threat detection framework that integrates standalone behavioral traits with sequential activity features through RNNs, thus improving detection accuracy compared to approaches that consider only one activity dimension. However, this method is highly dependent on predefined feature extraction and may struggle to adapt to unseen behavioral patterns.

Manoharan et al. \cite{hong2022graph} proposed a bi-channel insider threat detection framework that integrates features extracted from user logs with relational features derived from organizational graphs using GNNs, achieving improved detection performance on the CERT 4.2 dataset. While effective, their approach may face limitations in settings with sparse or dynamically changing organizational structures. Similarly, recent reviews \cite{Manoharan2024Review} highlight various insider threat detection methods, including user activity monitoring, network traffic analysis, and machine learning-based anomaly detection, underscoring the need to consider both individual behaviors and contextual interactions. Nevertheless, most traditional methods focus primarily on standalone user activities, often neglecting multi-dimensional dependencies that could further enhance detection accuracy.

However, all of the above graph-based works either use an explicit graph or an implicit graph, which may be suboptimal, as mentioned in the introduction. To counter the loss of information due to using either implicit or explicit graphs, we propose using both an explicit graph constructed with predefined rules and an implicit graph learned during training. This combined approach aims to leverage the strengths of both graph types, ensuring a comprehensive representation and more effective anomaly detection.

\section{Methodology}

Our proposed method comprises four main components: (1) Activity Encoding, (2) Log Feature extraction, (3) Graph Construction, (4) Node embedding generation and attention (5) Anomaly detection.

\begin{figure*}
    \centering
\includegraphics[width=0.9\linewidth]{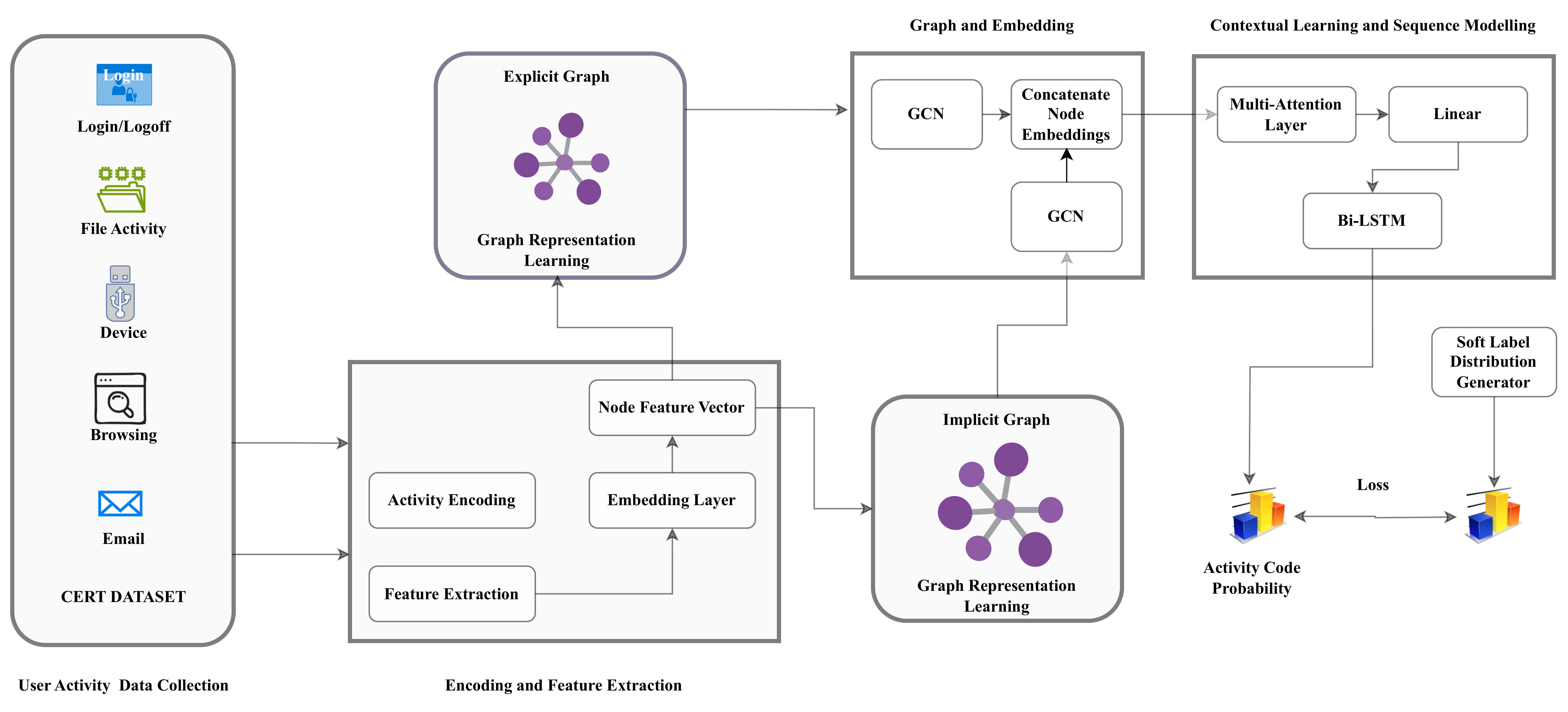}
 \captionsetup{justification=centering} 
  \caption{Overall Insider Detection Architecture.}
    \label{fig:insider-archi}
\end{figure*}

The activity encoding step assigns a numeric token to each user activity by combining its type and timestamp into a unique code, integrating activity type and occurrence time slot on. In the log feature extraction step, features indicative of activities from various events, such as the login event, login time, assigned PC, or supervisor PC, etc., are extracted. Following this, we construct an explicit graph based on two explicit rules and an implicit graph learned during training. Two GCNs are used to generate node embedding for both the explicit graph and implicit graph, which are concatenated and passed through a Multi-Head attention layer. The output of each head is then concatenated and passes through a final linear layer. Finally, the attention-refined node embedding is passed through a Bi-LSTM to predict the probability of each activity code. In anomaly detection, if the predicted probability of the mask activity is lower than a threshold, it is flagged as anomalous. The overall architecture is shown in Figure \ref{fig:insider-archi}.

\subsection{Activity Encoding}

Inspired by the work in \cite{cai2024lan}, we encode the different types of user activities into numerical values by applying a label encoder to the array of activity types, which includes 'Logon', 'Logoff', 'email', 'Http', 'File Open', 'File Write', 'Connect', and 'Disconnect'. This process transforms these activity types into corresponding numerical identifiers.

Next, we construct a unique activity code for each user activity by combining the encoded activity type with 
the specific hour during which the activity occurred. Formally, let \( \text{type}(a_i) \) represent the encoded 
activity type ID of activity \( a_i \), and \( \text{time}(a_i) \) denote the hour value of the activity's 
timestamp. The activity code \( c_i \) is then computed as:

\begin{equation}
    c_i = \text{type}(a_i) \times 24 + \text{time}(a_i). 
\end{equation}
This method integrates both the type and timing of an activity into a single numeric value, making it easier to handle and analyze within machine learning models.

\subsection{Log Feature Extraction}\label{log_feature_extraction}
From the individual log entries of each user, feature extraction is performed to create a feature vector for each activity. The feature extracted for each activity varies according to the log type. The features extracted for each log type are as follows:

\begin{enumerate}
    \item Logon/Logoff: Supervisor PC access  (binary  feature), assign PC access (binary feature), the elapsed time after the normal working hour (Normal working hours considered (9 am to 5 pm), PC access on the weekends (binary feature)
    \item File: Supervisor PC Access (binary feature), assign PC access (binary feature), the elapsed time after normal working hours, PC access on the weekend (binary feature), File transferred to removable (binary feature), File transferred from removable (binary feature), File open (binary feature), File write (binary feature), File delete (binary feature), File Copy (binary feature), Compressed file type (binary feature), Image File Type (binary feature), Document File Type (binary feature), Text file type (binary feature), Executable file type (binary feature), Other file types (binary feature), flag word count in file content (List of flag words: Jobs, Key-logging, key log)
    \item Device: Supervisor PC Access (binary feature), assign PC access (binary feature), the elapsed time after the normal working hours (Normal working hours considered (9 am to 5 pm), PC access on the weekend (binary feature), Device connected(binary feature), Device disconnected (binary feature)
    \item Email: Supervisor PC Access (binary feature), assign PC access (binary feature),
    elapsed time after the normal working hours (Normal working hours considered (9 am to 5 pm), PC access on the weekends (binary feature), number of email addresses outside the organisation, number of email addresses inside the organisation, number of email address BCC outside the organisation, number of email address BCC inside the organisation, number of email address cc outside the organisation, number of email address cc inside the organisation, aggregation of the file size of compressed file type attached with the email, aggregation of the file size of image file type attached with the email, aggregation of the file size of document file type attached with the email, aggregation of the file size of text file type attached with the email, aggregation of the file size of executable type attached with the email, aggregation of the file size of other file type attached with the email, flag word count in file content( List of flag words: Jobs, Key-logging, key log).
    \item HTTP: Supervisor PC access (binary feature), assign PC access (binary feature), the elapsed time after normal working hours (Normal working hours considered (9 am to 5 pm), PC access on weekends (binary feature),
    flag URL access (Wikileaks.org, Dropbox.com), flag word count in file content (List of flag words: Jobs, Key-logging, key log), visit activity (binary feature), download activity (binary feature), upload activity
    (binary feature).
\end{enumerate}

\subsection{Graph Construction}\label{graph_construction}

User activities are divided into unique sessions, where a login marks the start of the session, and a logoff marks the end. If logs occur without a corresponding logoff, the following login marks the end of the previous session. This segmentation organizes activities into coherent, time-bound sessions. Sessions are further divided into sub-sessions, each containing up to \( n=50 \) activities. This division manages longer sessions by breaking them down into smaller, more manageable chunks. Sub-sessions with fewer than five activities merge with adjacent sub-sessions to ensure sufficient data density, preventing sparsity issues that could hinder analysis.

For each sub-session, sequences are generated by systematically masking the next activity at each time step. Specifically, for each activity \( a_t \) at time step \( t \), a masked sequence \( S' = \{a_1, \ldots, a_{t-1},  \langle MASK \rangle, a_{t+1}, \ldots, a_n\} \) is created. Masking for intrusion detection was first reported in LAN (\cite{cai2024lan}), although this paper manages the sub-sessions differently. These masked sequences and their labels (normal activity or insider activity) are stored for subsequent analysis.

\subsubsection{Explicit Graph Construction}

In constructing the explicit graph for our post-intrusion detection framework, we employ a systematic approach to capture both the sequential and type-based relationships between user activities. The primary goal is to create a graph that effectively represents the structure and interactions within the sequence of activities. The process begins by defining the nodes and edges of the graph based on the sequence of user activities.

Each node in the graph corresponds to an individual activity $c_i$ within the sequence, and the edges represent the relationships between these activities. We utilize two primary rules to establish these connections: sequential connections and type-based connections (edges). Firstly, we establish \textbf{sequential connections} between activities. This involves connecting each activity to the one immediately following it, as well as the one immediately preceding it. These connections ensure that the temporal order of activities is preserved in the graph structure. Specifically, for a sequence of \( n \) activities, edges are created between node \( i \) and node \( i+1 \), and between node \( i+1 \) and node \( i \), for all \( i \) from 0 to \( n-1 \). This bidirectional linking captures the immediate temporal dependencies between consecutive activities, ensuring that the sequence's flow is maintained in the graph. In addition to the sequential connections, we also incorporate \textbf{type-based connections} to capture similarities based on activity types. To achieve this, we establish connections between activities of the same type (email–email, logon–logoff, connect–disconnect, http–http, file – open, file – delete, and file–write).

\subsubsection{Implicit Graph Construction}\label{sub-sec:graph-const}

In our approach to constructing implicit graphs, we focus on dynamically learning the graph structure from the node embeddings derived from feature vectors. This process is crucial for capturing the latent relationships between nodes that may not be explicitly observable. The process begins by obtaining the node embeddings $E$ from the feature vectors by passing each activity in the sub-sequence $S'$ through an embedding layer. Using the embedding, a similarity matrix is calculated by performing matrix multiplication between the node embeddings and their transpose, followed by normalizing this matrix to make the similarity scores proportional and comparable across different node pairs. This normalization is done using the formula:

\begin{equation}
\theta = \frac{E E^T}{\|E\| \|E^T\| + \epsilon}.
\end{equation}

where \(E\) is the matrix of node embeddings, and \(\epsilon\) is a small constant to prevent division by zero.

To generate the regularised adjacency graph we utilized the Gumbel softmax trick in \cite{yu2022regularized}. The grumble softmax trick discretizes the continuous similarity matrix into a binary adjacency matrix while maintaining the graph's differentiability for gradient descent optimization. The process of graph generation through Gumble Softmax is also considered a regularized graph generation because of the probabilistic nature of the Gumbel Softmax, which introduces randomness in the selection of edges and helps in learning more generalized graph structures. It effectively controls the density of the graph, reducing overfitting by preventing overly dense connections that might not be representative of true underlying relationships.

To generate a sparse implicit adjacency matrix \( A_\text{implicit} \) from \( \theta \) the following equation is employed:

\begin{equation}
    A_\text{implicit} = \sigma\left(\frac{\log(\frac{\theta_{ij}}{1 - \theta_{ij}}) + g_1 - g_2}{s}\right).
\end{equation}

In this formula, \( g_1 \) and \( g_2 \) are Gumbel noise terms, \( s \) is the temperature parameter, and \( \sigma \) is the logistic sigmoid function. This step effectively discretizes the continuous probabilities into a binary adjacency matrix, representing the graph's edges.

\subsection{Node embedding generation and attention}

Two separate Graph Convolutional Networks (GCNs), one for the explicit graph and another for the implicit graph, are utilized to generate the node embedding of both the explicit and implicit graphs. 
First, the initial node embeddings are generated from the input feature vectors using a linear transformation:

\begin{equation}
    E = W_{emb} X + b_{emb} .
\end{equation}
where \( X \) is the input feature matrix, \( W_{emb} \) is the weight matrix, \( b_{emb} \) is the bias vector, and \( E \) represents the initial node embeddings. The node embeddings corresponding to the explicit graph are passed through the GCN layer.

\begin{equation}
    H_\text{explicit}^{(n)} = \text{ReLU}(A_\text{explicit} H_n W_{explicit}^{(n)}).
\end{equation}

Where \( H_0 = E \) is the initial embedding, \( W_{explicit}^{(n)} \) is the weight matrices for the explicit GCN layers and $n = 1, 2, \dots, N$ denotes the layer index. Simultaneously, another set of GCN layers processes the embeddings using the learned (implicit) graph adjacency matrix \( A_\text{implicit} \):

\begin{equation}
    H_\text{implicit}^{(n)} = \text{ReLU}(A_\text{implicit} H_n W_{implicit}^{(n)}).
\end{equation}

The outputs from both GCN pathways are concatenated then to integrate the information from both graphs:
\begin{equation}
C = [H_\text{explicit}^{(n)} \parallel H_\text{implicit}^{(n)}]. 
\end{equation}

We then compute the query, key, and value matrices for the multi-head attention mechanism:
\begin{equation}
Q = CW_Q + b_Q .
\end{equation}
\begin{equation}
K = CW_K + b_K .  
\end{equation}
\begin{equation}
V = CW_V + b_V .
\end{equation}

Where \( W_Q, W_K, W_V \) are the weight matrices, and \( b_Q, b_K, b_V \) are the bias vectors for the query, key, and value transformations, respectively.

The multi-head attention mechanism is applied to focus on the most relevant features:
\begin{equation}
    \text{Attention}(Q, K, V) = \text{softmax}\left(\frac{QK^T}{\sqrt{d_k}}\right)V.
\end{equation}  
Where \( d_k \) is the dimension of the key vectors. For multi-head attention, the outputs of multiple attention heads are concatenated and linearly transformed:
\[ \text{MultiHead}(Q, K, V) = \text{Concat}(\text{head}_1, \ldots, \text{head}_h) W^O \],
with
\[ \text{head}_i = \text{Attention}(Q W_i^Q, K W_i^K, V W_i^V) \].
where \( W_i^Q, W_i^K, W_i^V \) are the projection matrices for the \( i \)-th head, and \( W^O \) is the output projection matrix. The result from the multi-head attention mechanism, denoted as $Attn$ is passed through a linear layer to get the final node embedding.
\begin{equation} 
O = A W_O + b_O. 
\end{equation}

\subsection{Anomaly Detection}\label{Anomaly Detection}

In our anomaly detection framework, we utilize the node embeddings generated by our dual GCN and attention mechanism to predict the probability distribution of all possible activity codes given the subsequence $S'$. After generating the node embeddings, we organize them temporally. Given that activity sequences can vary in length, we pad them to a uniform length to maintain consistency during processing. The padded sequences are then fed into a bidirectional Long Short-Term Memory (Bi-LSTM) network. The Bi-LSTM is essential for capturing temporal dependencies from both past and future contexts, providing a comprehensive understanding of the sequence dynamics. Mathematically, the Bi-LSTM consists of two LSTMs: one processing the sequence forward (\( \overrightarrow{h_t} \)) and the other backwards (\( \overleftarrow{h_t} \)). For an input sequence \( x = (x_1, x_2, \ldots, x_T) \), the forward and backward LSTMs compute the hidden states as follows:

The forget gate \( f_t \) controls which parts of the previous cell state \( c_{t-1} \) should be retained or discarded. It is defined as:

\begin{equation}
 f_t = \sigma(W_f x_t + U_f h_{t-1} + b_f).   
\end{equation}

The input gate \( i_t \) determines which new information will be added to the cell state. It is calculated as:

\begin{equation}
   i_t = \sigma(W_i x_t + U_i h_{t-1} + b_i). 
\end{equation}

Next, the candidate cell state \( \tilde{c}_t \) represents the potential updates to the cell state based on the current input and the previous hidden state:

\begin{equation}
\tilde{c}_t = \tanh(W_c x_t + U_c h_{t-1} + b_c).
\end{equation}

The actual cell state \( c_t \) is updated by combining the previous cell state, scaled by the forget gate, and the candidate cell state, scaled by the input gate:

\begin{equation}
  c_t = f_t \odot c_{t-1} + i_t \odot \tilde{c}_t.  
\end{equation}

The output gate \( o_t \) determines which parts of the cell state will contribute to the next hidden state:

\begin{equation}
o_t = \sigma(W_o x_t + U_o h_{t-1} + b_o).    
\end{equation}

Finally, the hidden state \( h_t \) is computed by applying the output gate to the updated cell state, using a non-linear activation function:
\begin{equation}
 h_t = o_t \odot \tanh(c_t). 
\end{equation}

\begin{equation}
    \overrightarrow{h_t} = \text{LSTM}(x_t, \overrightarrow{h_{t-1}}).
\end{equation}

\begin{equation}
\overleftarrow{h_t} = \text{LSTM}(x_t, \overleftarrow{h_{t+1}}).
\end{equation}

The final hidden state at each time step \( t \) is the concatenation of the forward and backward hidden states:

\begin{equation}
h_t = [\overrightarrow{h_t}; \overleftarrow{h_t}].
\end{equation}

The Bi-LSTM output is aggregated, typically by averaging the hidden states across all time steps, to form a single representation for each sequence:

\begin{equation}
    h_{\text{agg}} = \frac{1}{T} \sum_{t=1}^{T} h_t.
\end{equation}

This aggregated representation is then passed through a fully connected layer to predict the probability distribution over all possible activity codes:

\begin{equation}
    y = \text{softmax}(W_o h_{\text{agg}} + b_o).
\end{equation}

Here, \( W_o \) and \( b_o \) are the weight matrix and bias vector for the output layer. To identify anomalies, we focus on the probability assigned to the masked activity code. If this probability is lower than a predefined threshold, the activity is flagged as anomalous. By combining relational data from the GCNs with sequential patterns from the Bi-LSTM, our model robustly identifies subtle anomalies indicative of insider threats, leveraging the strengths of both graph-based and temporal modelling techniques. To train the whole architecture end to end, we use the loss in \cite{cai2024lan}. This loss function integrates both self-supervised learning from normal activities and supervised learning from abnormal activities. Given an activity sequence \( S = \{a_1, a_2, \ldots, a_n\} \) with corresponding anomaly labels \( q \in \{0, 1\}^n \), where \( q_i = 1 \) indicates an abnormal activity. For abnormal activities, we create a new soft label distribution is created using the following equation:
\begin{equation}
Y' = Y \odot (1 - \Omega) + \frac{1}{M-1} \Omega \odot (1 - Y \odot \Omega).    
\end{equation}

where \( \Omega = q \otimes \mathbf{1}_M \), and \( \mathbf{1}_M \) is a vector of ones with length \( M \). \( \odot \) denotes the Hadamard product. \( Y \in \mathbb{R}^{n \times M} \) represents the one-hot encoded labels of the activity sequence \( S \).    For abnormal activities (\( q_i = 1 \)):
\begin{equation}
  Y'_{ij} = 
   \begin{cases} 
   0 & \text{if } j = c_i , \\
   \frac{1}{M-1} & \text{if } j \neq c_i .
   \end{cases}
\end{equation}

To counter the imbalance during training, we oversample the abnormal sequence $r$ times, which is the imbalance ratio. The following equation gives the final loss. 

\begin{equation}
  L = - \frac{1}{n-1}\sum_{j=1}^{M} Y'_{ij} \log(Y_{ij}) . 
\end{equation}

To detect anomalies, we compare the predicted probability of the masked activity code to a predefined threshold. If the probability \( y_m \) assigned to the masked activity code \( c_m \) is lower than the threshold \( \tau \), the activity is flagged as anomalous:

\[
\text{Anomaly} = 
\begin{cases} 
\text{True} & \text{if } y_m < \tau ,\\
\text{False} & \text{if } y_m \geq \tau .
\end{cases}
\]

This condition ensures that mask activity significantly deviating from expected behaviour is detected, leveraging the comprehensive modelling capabilities of the combined GCN, attention mechanism, and Bi-LSTM architecture.

\section{Experiments}

\subsection{Dataset}

In the experiments, we use the CERT insider threat detection dataset by Carnegie Mellon University, specifically the CERT r5.2 dataset (10.37 GB) and the latest CERT dataset r6.2 (22.18 GB) \cite{glasser2013bridging}. These datasets include user behaviours across various actions, such as logging on and off, sending emails, accessing files, connecting devices, and more. Each dataset version typically comes with different user scenarios, including non-malicious and malicious actions, to aid in developing and testing insider threat detection systems. The datasets contain multiple users with distinct roles (e.g., employees, contractors) and behavioural patterns, with certain users exhibiting "insider threat" behaviours. They include comprehensive information on synthetic insider threat activities, with various event types, including logon/logoff actions, file access, emails, and device connections. The datasets are typically provided as multiple files, each representing different activities or system logs. Common files include: Logon.csv, Email.csv, File.csv, HTTP.csv, Device.csv and Psychometric.csv. 

The structure and features of the CERT insider threat datasets make them well-suited for developing insider threat detection models, providing a realistic simulation of workplace activities with various potential insider threats. Due to the size of the dataset, we randomly select ten users from the r5.2 dataset and include all 29 insiders from the r6.2 dataset. Following the approach in \cite{cai2024lan}, we use the user activity data from 2010 for training and validation and the data from January 2011 to June 2011 for evaluating all methods for the r5.2 dataset. For r6.2, we select data from July to September 2010 for all insider users except CDE1846, CMP2946, and MBG3183. For users CDE1846 and CMP2946, we include data from January 2011 to April 2011. For user MBG3183, we consider data from October to December 2010. These periods contain all the insider activity in the dataset. After filtering the r6.2 data, we split it into 70\% for training and 30\% for testing.  Since each sub-session contains up to 50 activities and masking is done sequentially, each sub-session will result in 50 masked sequences. Consequently, each sub-session will have 50 corresponding graphs. 

For r5.2, the training dataset contains 146,721 normal masked sub-sequences that do not include any insider activity and 84 masked sub-sequences with malicious insider activity. During testing, there are 17,429 normal masked sub-sequences without insider activity and 33 masked sub-sequences with insider activity. For r6.2, the training dataset includes 172,970 normal masked sub-sequences without any insider activity and 339 masked sub-sequences with malicious insider activity. In the testing phase, there are 74,145 normal masked sub-sequences without insider activity and 131 masked sub-sequences with insider activity. Finally, to address the imbalance, the malicious dataset is replicated according to the imbalance ratio
 \( \frac{N_{\text{maj}}}{N_{\text{min}}} \), creating a balanced training set.

\subsection{Implementation Details}
First, we assign an hour ID to each user's activity. For example, if an activity occurs at 6:10 AM, the hour ID assigned is 6. This grouping helps us organise the activities into hourly intervals. Next, we extract features for each activity as described in our feature extraction process in section \ref{log_feature_extraction}. These features encapsulate important attributes and behaviours associated with each activity. The experiment aggregates activity features with the same activity code that occur within the same hour on the same date to smooth out variation in the feature. The features are scaled, transforming each to a distribution centred around 0 with a standard deviation of 1 for better convergence. Activities within the same hour are split into sub-sessions of at most 50 activities. If an hour has fewer than five activities, it combines with the next hour. To address the dataset imbalance in the CERT dataset, instances of malicious activity are replicated by a factor corresponding to the calculated imbalance ratio. This augmentation aims to ensure a more equitable representation of both malicious and non-malicious instances for effective training.

Using the sequence described in section \ref{graph_construction} (Explicit Graph Construction), we construct the explicit graph. In contrast, the implicit graph is dynamically generated during training using node embeddings and the Gumbel softmax trick as outlined in section \ref{graph_construction} (Implicit Graph Construction). Each node's features in the graph first pass through an embedding layer sized 54x54. We used two different GCNs for explicit and implicit graphs to generate the node embedding, which we termed DualGCNWithAttention. The DualGCNWithAttention model utilizes this embedded input dimension of 54, initializing two distinct Graph Convolutional Networks (GCNs) tailored for explicit and implicit graph structures. Each GCN starts with an input layer sized 54x16, where 54 denotes the number of input features, and 16 represents the dimensionality of the hidden layer. Additional hidden layers sized 16x16 further refine the outputs from these GCNs. Following GCN processing, the outputs from both explicit and implicit GCNs are combined and processed through a multi-head attention mechanism comprising eight heads. This mechanism is pivotal in capturing comprehensive feature interactions and dependencies across the aggregated outputs. The resultant output from the attention mechanism then undergoes further refinement through a linear output layer sized 16x16, producing the final node embeddings ready for subsequent prediction tasks.

Since batch learning is employed, graphs from different subsequences are combined into one big graph and processed in parallel. Therefore, after the output from DualGCNWithAttention, we segment the node embedding using the batch number so that each subsequence has independent node embeddings. These node embeddings are further processed to ensure uniformity in sequence length across batches. Specifically, sequences of node embeddings within each batch are padded to match the maximum sequence length. The padded node node embeddings are then used as input to a Bidirectional LSTM (Bi-LSTM). The Bi-LSTM is designed to predict the probabilities of 192 different activity codes based on the node embeddings. The Bi-LSTM initializes with an input dimension of 16. The LSTM comprises two layers, utilizing a hidden dimension of 32 for each direction (forward and backwards). Due to its bidirectional nature, the output dimension of the LSTM is doubled, resulting in a hidden dimension size of 64. Subsequently, the final output layer is configured with a size of 192, matching the number of distinct activity codes for which the model predicts probabilities. Both the model DualGCNWithAttention and Bi-LSTM are trained in an end-to-end manner using the loss described in section \ref{Anomaly Detection}. The experiments utilize Adam as an optimizer with a learning rate of 0.0001.

\subsection{Experimental Results and Analysis}

To evaluate the performance of our model, we adopt metrics such as Area Under the Curve (AUC), Detection Rate (DR), and False Positive Rate (FPR), similar to methodologies employed in previous studies (\cite{Log2vec},\cite{cai2024lan}). The Area Under the Curve (AUC) serves as a primary indicator, capturing the model's ability to discriminate effectively across a range of decision thresholds. Detection Rate (DR), also known as True Positive Rate (TPR), measures the percentage of actual positive instances correctly identified by the model out of all actual positives. Finally,  False Positive Rate calculates the proportion of negative instances incorrectly classified as positive by the model. This metric helps assess the model's specificity in distinguishing between normal and abnormal activities. The formulae for AUC, DR and FPR are as follows:

 \begin{equation}
 \text{AUC} = \text{Area Under ROC Curve}. 
 \end{equation}
 \begin{equation}
 \text{DR/TPR} = \frac{\text{TP}}{\text{TP} + \text{FN}}  .
 \end{equation}
 \begin{equation}
\text{FPR} = \frac{\text{FP}}{\text{FP} + \text{TN}} .\
 \end{equation}

\begin{figure}
\includegraphics[width=10cm]{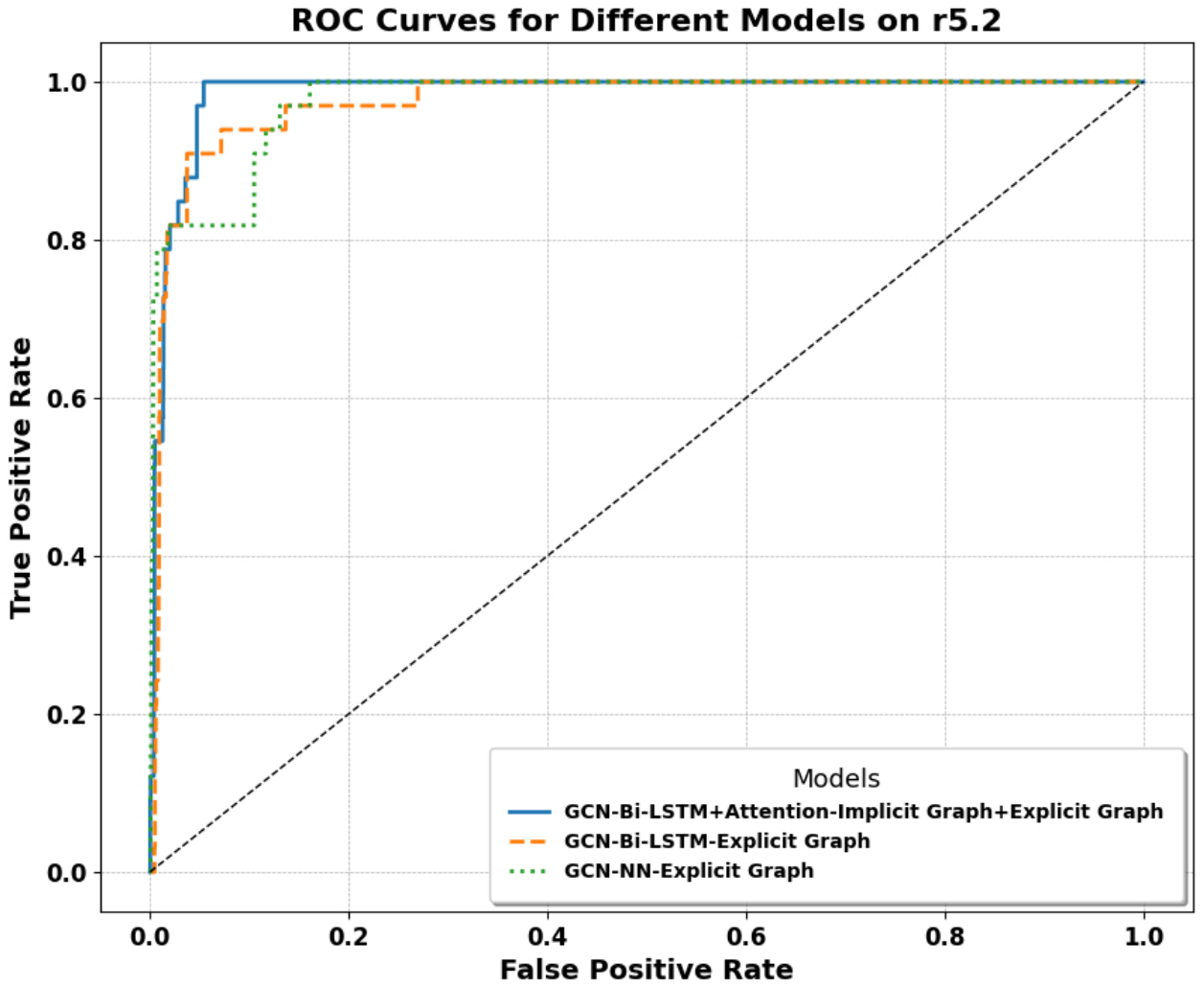}
 \caption{Comparison of ROC Curves Between Different Models on the r5.2 CERT Dataset: Evaluating Performance with Explicit Graphs Versus Combined Explicit and Implicit Graphs.}
 \label{fig:r5.2_roc}
 \includegraphics[width=10cm]{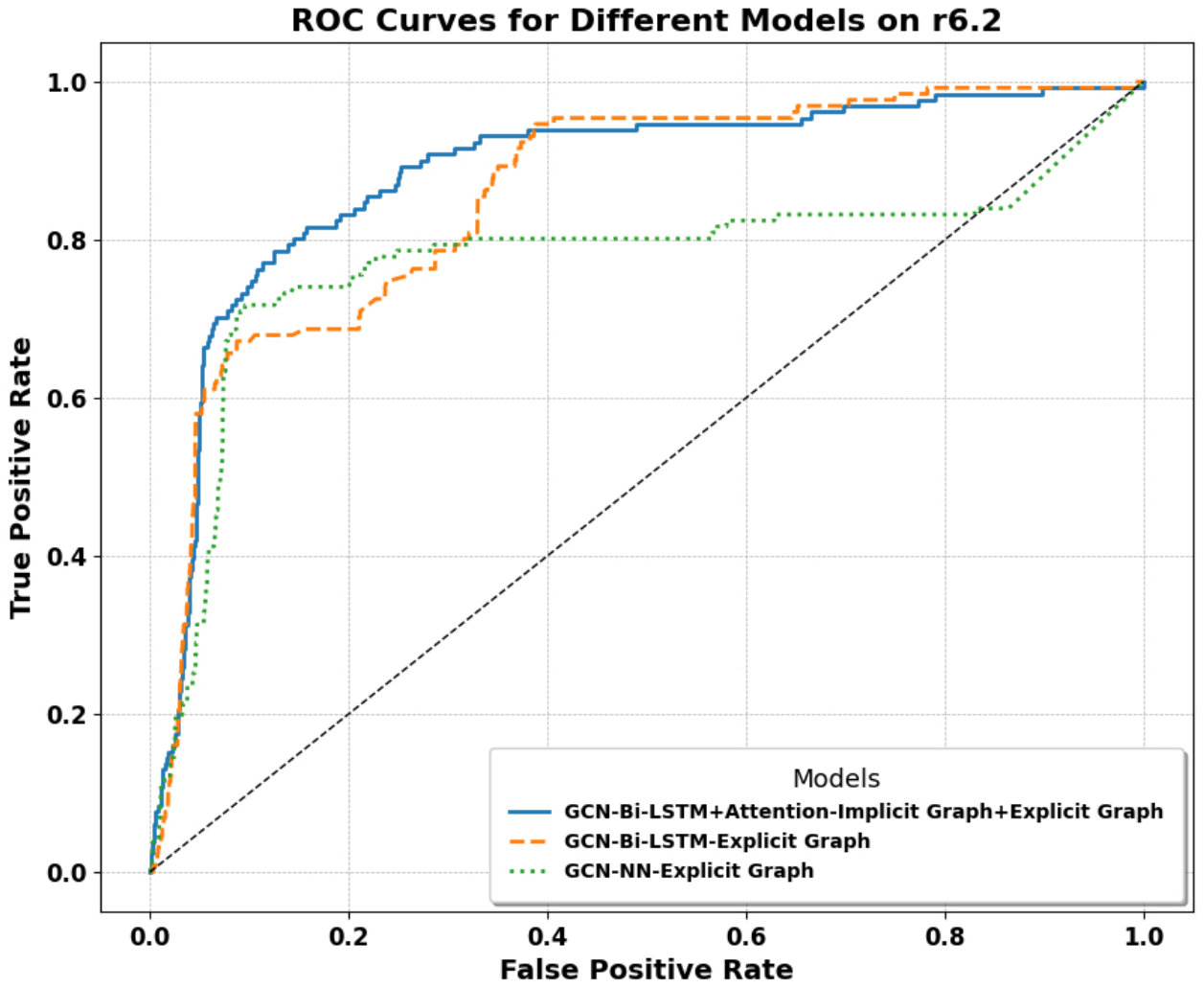}
 \caption{Comparison of ROC Curves Between Different Models on the r6.2 CERT Dataset: Evaluating Performance with Explicit Graphs Versus Combined Explicit and Implicit Graphs.}
 \label{fig:r6.2_roc}
\end{figure}

 Figure \ref{fig:r5.2_roc} and \ref{fig:r6.2_roc} show the ROC curve on the two datasets r5.2 and r6.2, respectively. The ROC curve is a graphical representation of a model's performance to assess the performance of a binary classification model, illustrating the trade-off between the true positive rate and the false positive rate. The model that combines a Graph Convolutional Network (GCN) with Bidirectional Long Short-Term Memory (Bi-LSTM) and an attention mechanism, utilizing both implicit and explicit graph representations, achieves the highest performance, as indicated by its ROC curve, which is closest to the top-left corner of the plot. This suggests that combining both implicit and explicit graph representations enhances the model's ability to distinguish between positive and negative classes effectively on both datasets. The model that combines GCN with the Bi-LSTM component, while slightly lower than the first, performs better than the one using GCN and Neural Network (NN) combination on both the datasets, indicating that the Bi-LSTM component effectively captures sequential dependencies within the explicit graph structure. The inclusion of implicit graph representations alongside explicit ones significantly enhances model performance, as evidenced by the consistently higher ROC curves of the GCN-Bi-LSTM+Attention model. However, the r6.2 dataset presents a more challenging classification task, as reflected in the lower AUC values across all models.

 In practical application, we need to set a threshold for detection. In many practical scenarios, a specific FPR is targeted based on application requirements. For the r5.2 dataset, a target FPR of 0.05 was chosen, reflecting the better ROC performance of the models on this dataset, and for the r6.2 dataset, a target FPR of 0.09 was set due to the relatively lower ROC performance observed for the models on this dataset. More specifically, to find the optimal threshold, we look for the highest threshold where the FPR is less than or equal to the target value \( \text{FPR}_{\text{target}} \). Mathematically, we seek the threshold \( t_{\text{optimal}} \) such that:


\begin{equation}
   \text{FPR}(t_{\text{optimal}}) \leq \text{FPR}_{\text{target}}.
\end{equation}

\begin{equation}
    t_{\text{optimal}} = \max\{ t \, | \, \text{FPR}(t) \leq \text{FPR}_{\text{target}} \} .  
\end{equation}

 We find the threshold \( t \) where the FPR first meets or drops below the target value and is the largest such threshold. The results obtained are shown in Table \ref{tab:cert_performance_table}. The r5.2 dataset results indicate high performance across all models, with particularly noteworthy outcomes for the GCN-Bi-LSTM+Attention-Implicit Graph+Explicit Graph model. This model achieves an impressive AUC of 98.62, suggesting it is highly effective in distinguishing between the positive and negative classes. Additionally, it has a perfect DR/TPR of 100\%, meaning it correctly identifies all positive instances without missing any. The FPR of 0.05 indicates that only 5\% of negative instances are incorrectly classified as positive.

\begin{table}
\centering
\begin{threeparttable}
\caption{Performance metrics for different models and datasets.}
\label{tab:cert_performance_table}
\begin{tabular}{|c|c|c|c|c|}
\hline
\textbf{Dataset} & \textbf{Model} & \textbf{AUC} & \textbf{DR/TPR} & \textbf{FPR} \\
\hline
\multirow{3}{*}{CERT r5.2} 
 & GCN-BiLSTM-Explicit & 97.47 & 93.93 & 0.09\\
\cline{2-5}
 & GCN-BiLSTM+Attn-Im+Ex & 98.62 & 100 & 0.05  \\
 \cline{2-5}
 & GCN-NN-Explicit & 97.56 & 81.81 & 0.05 \\
 \cline{2-5}
& GCN-NN-Implicit & 94.81 & 77.15 & 0.07 \\
\cline{2-5}
& GCN-NN-Im+Ex & 96.48 & 88.10 & 0.06 \\
\cline{2-5}
& GCN-NN  & 92.15 & 85.37 & 0.05 \\
\hline
\multirow{6}{*}{CERT r6.2} 
 & GCN-BiLSTM-Explicit  & 85.6 & 67.17 & 0.14\\
\cline{2-5}
 & GCN-BiLSTM+Attn-Im+Ex & 88.48 & 80.15 & 0.15 \\
 \cline{2-5}
 & GCN-NN-Explicit & 77.27 & 74.51 & 0.14 \\
 \cline{2-5}
& GCN-NN-Implicit & 78.83 & 70.12 & 0.17 \\
\cline{2-5}
& GCN-NN-Im+Ex & 87.10 & 76.93 & 0.14 \\
\cline{2-5}
& GCN-NN  & 77.28 & 71.80 & 0.096 \\
\hline
\end{tabular}
\begin{tablenotes}
\footnotesize
\item GCN: Graph Convolutional Network; NN: Neural Network; BiLSTM: Bidirectional Long Short-Term Memory; Attn: Attention; Im: Implicit Graph; Ex: Explicit Graph.
\end{tablenotes}
\end{threeparttable}

\end{table}

\begin{table}
\centering
\caption{Comparison with state-of-the-art methods.}
\begin{tabular}{|l|c|c|c|c|c|}
\hline
\textbf{Model}& \textbf{Dataset} & \textbf{AUC} & \textbf{TPR} & \textbf{FPR} \\
\hline
Deep Log & CERT r5.2 & 86.41 & 81.89 & 0.19 \\
\hline
LAN & CERT r5.2 & 97.07 & 96.78 & 0.04 \\
\hline
*Our work & CERT r5.2 & 98.62 & 100 & 0.05 \\
\hline
Log2-vec & CERT r6.2  & 86 & 79.16 & 0.18 \\
\hline
*Our work & CERT r6.2 & 88.48 & 80.15 & 0.15 \\
\hline
\end{tabular}

\label{tab:cert_comparision}
\end{table}
 
In contrast, the GCN-NN-Explicit Graph model, while still performing well, has a slightly lower AUC of 97.56 and a DR/TPR of 81.81\%. This indicates it misses approximately 18.19\% of positive cases. However, it maintains a low FPR of 0.05, matching the highest-performing model in this metric. The GCN-Bi-LSTM-Explicit Graph model strikes a balance with an AUC of 97.47 and a higher DR/TPR of 93.93\% but at the cost of a slightly higher FPR of 0.09. This model is more likely to identify positive cases correctly than the GCN-NN-Explicit Graph model, but it does so with a slight increase in false positives.

 The r6.2 dataset presents a more challenging scenario for all models, as indicated by generally lower performance metrics compared to the r5.2 dataset. The GCN-Bi-LSTM+Attention-Implicit Graph+Explicit Graph model again leads with the highest AUC of 88.48, demonstrating its superior ability to differentiate between classes even in more challenging conditions. It achieves a DR/TPR of 80.15\%, meaning it correctly identifies 80.15\% of positive cases but with an increased FPR of 0.15. This suggests that while effective, the model also incorrectly classifies 15\% of negative cases as positive. The GCN-NN-Explicit Graph model has the lowest performance on the r6.2 dataset, with an AUC of 77.27 and a DR/TPR of 74.51\%. The TPR obtained indicates significant challenges in accurately identifying positive cases, compounded by a higher FPR of 0.14. This model struggles the most in distinguishing between positive and negative instances in this dataset. The GCN-Bi-LSTM-Explicit Graph model performs moderately with an AUC of 85.6 and a DR/TPR of 67.17\%. 

 The comparative analysis of the r5.2 and r6.2 datasets shows the variability in model performance based on dataset characteristics. The GCN-Bi-LSTM+Attention-Implicit Graph+Explicit Graph model consistently performs the best across both datasets, achieving the highest AUC and DR/TPR values. This model's ability to leverage both implicit and explicit graph representations, along with sequential learning mechanisms, likely contributes to its superior performance. The findings suggest that while simpler models like the GCN-NN-Explicit Graph can perform adequately in less challenging scenarios (r5.2), they struggle with more complex datasets (r6.2). Conversely, more sophisticated models incorporating advanced techniques such as attention mechanisms and multi-graph representations offer better performance but may still face challenges with higher false positive rates in more complex contexts.

 In comparison with the state-of-the-art method shown in Table \ref{tab:cert_comparision}, our methods perform better in both datasets. DeepLog utilizes Long Short-Term Memory (LSTM) networks to predict whether each log entry is anomalous. On the r5.2 dataset, DeepLog achieves an AUC of 86.41, a TPR of 81.89\%, and an FPR of 0.19. While it demonstrates a reasonable capability in detecting anomalies, its performance is lower compared to other methods. LAN combines graph convolutional networks (GCNs), attention mechanisms, and LSTMs and uses an implicit graph to predict anomalous log entries. On the r5.2 dataset, LAN achieves an impressive AUC of 97.07, a TPR of 96.78\%, and an FPR of 0.04. On the r5.2 dataset, our model achieves the highest AUC of 98.62, a perfect TPR of 100\%, and an FPR of 0.05, outperforming other methods. Log2-vec creates a explicit heterogeneous graph and uses random walks to generate node embeddings for anomaly detection. On the r6.2 dataset, Log2-vec achieves an AUC of 86, a TPR of 79.16\%, and an FPR of 0.18. On the more challenging r6.2 dataset, our model also outperforms other methods with an AUC of 88.48, a TPR of 80.15\%, and an FPR of 0.15. Although the performance metrics are slightly lower on this more difficult dataset, our model still achieves the best results among the compared methods.

 Explicit and implicit graph construction can be computationally intensive, particularly in the context of large-scale insider threat detection. In order to reduce the computational requirement, future research should investigate methods to streamline the construction process by employing pruning and sampling techniques to reduce the graph's size without compromising detection accuracy. Pruning could involve the removal of redundant or irrelevant nodes and edges that do not significantly contribute to threat analysis. For instance, routine interactions with low-sensitivity resources could be deprioritized, while high-risk activities remain integral to the graph. Additionally, sampling techniques can aid in managing graph complexity by selecting key nodes and edges. Node sampling could prioritize users engaged in high-risk behaviours, such as unusual access to sensitive resources, while less critical nodes are sampled at lower rates. Edge sampling could retain only the most significant connections, focusing on interactions indicative of malicious behaviour, such as accessing high-privilege s stems. Random walk sampling enables the system to concentrate on suspicious parts of the graph, starting from flagged nodes and exploring their immediate network, thereby reducing the need to process the entire graph. Lastly, temporal sampling could prioritize recent activities, ensuring that the system captures the most relevant actions in real-time while minimizing the burden of maintaining outdated data.

\section{some real insider threat examples}

 Insider threats continue to pose serious risks to organizations of all sizes and sectors, as shown by numerous high-profile cases. In 2023, Tesla revealed that two former employees leaked sensitive personal information, including names, addresses, and social security numbers of over 75,000 staff, along with customer banking details and production secrets, causing lasting reputational harm. At Yahoo in 2022, a research scientist downloaded about 570,000 pages of trade secrets related to its AdLearn product before moving to competitor The Trade Desk, leading to lawsuits for intellectual property theft. Not all incidents were malicious: in 2022, Microsoft employees inadvertently exposed login credentials for Azure systems on GitHub, which could have enabled attackers to access critical infrastructure if not caught by a security firm. Proofpoint also faced risks when a departing employee copied confidential sales data to benefit rival Abnormal Security, bypassing the company’s own data loss prevention tools. In 2020, Twitter (now X) employees fell victim to spearphishing that let hackers hijack 130 high-profile accounts for a bitcoin scam, highlighting the risks of social engineering. Earlier, in 2016, a Google engineer stole thousands of files from its self-driving car project before joining Uber, leading to lawsuits and financial losses, while Marriott’s 2020 vendor-related breach exposed the records of 339 million guests due to compromised employee credentials, resulting in an £18.4 million GDPR fine. Apple, in 2022, accused startup Rivos of poaching over 40 employees, some of whom allegedly exfiltrated gigabytes of proprietary SoC design data, threatening years of research investment. Boeing saw negligence in 2017 when an employee emailed a spreadsheet containing hidden personal data of 36,000 coworkers to his wife’s personal account. Similarly, in 2023, Reddit disclosed a breach when an employee was tricked by a phishing site, exposing databases of historic user credentials. In a more destructive example, during the early COVID-19 pandemic in 2020, a former executive at Stradis Healthcare used secret accounts to sabotage shipping data, delaying PPE deliveries to hospitals. Collectively, these cases show that insider threats can emerge from disgruntled staff, careless employees, or third-party access, with impacts ranging from intellectual property theft and competitive disadvantage to regulatory fines, operational disruption, and significant reputational damage \cite{simpson2025}.

\section{Ethical Considerations}

The insider threat detection techniques in this work were tested on the CERT datasets, which are synthetic datasets simulating insider activities in a controlled environment. While valuable for model benchmarking, deploying these techniques in real-world organizations raises ethical issues that need to be addressed, particularly in handling sensitive employee data such as login times, file access, and communications.

In a real organisation, unlike synthetic data, accurate employee data introduces significant privacy concerns. Future implementations should limit data collection to only essential information necessary for detection. Incorporating privacy-preserving methods like data anonymisation or differential privacy will help protect individual identities while maintaining detection efficacy. The CERT datasets, although helpful, may need to capture the diversity of behaviours in a real organisation fully. There is a heightened risk of bias when implementing this approach in real-world scenarios, as it may unfairly target certain user behaviours or demographic groups. Future systems need to be trained on more diverse and representative datasets that encompass the full spectrum of roles, departments, and demographics present in organisations. Utilising fairness-aware algorithms and regularly reviewing model outputs through audits are critical steps to guarantee fair treatment of all employee groups. Additionally, implementing insider threat detection in a real-world context can create a perception of excessive surveillance, which may erode employee trust. Future systems should focus on selective monitoring of high-risk behaviours rather than blanket surveillance. Transparent policies outlining the scope and purpose of monitoring, with the ability to contest unfair flags, can help maintain trust and fairness.

Implementing Role-Based Access Control (RBAC) and Multi-Factor Authentication (MFA) is essential for restricting system and data access exclusively to authorised personnel. Additionally, a comprehensive data retention policy must be established. This policy should clearly specify the specific duration for which data is to be stored, followed by secure deletion protocols to ensure the data's confidentiality and integrity post-retention period. Finally, any real-world insider threat detection system must comply with legal frameworks such as the General Data Protection Regulation (GDPR), the California Consumer Privacy Act (CCPA), and other relevant data protection laws. Ensuring legal compliance not only protects the organisation but also fosters trust by respecting employee rights related to data access, storage, and deletion.

\section{Conclusion}
This study demonstrates that combining explicit and implicit graph representations with neural network architectures, such as Graph Convolutional Networks (GCNs), attention mechanisms, and Long Short-Term Memory (LSTM) networks, significantly enhances anomaly detection performance. This integrative approach leverages the unique strengths of each component, yielding superior results across diverse datasets. The explicit and implicit graph representations provide a dual approach to capturing relationships within the data. Explicit graphs offer clear, predefined connections, which are essential for understanding direct relationships. Implicit graphs, on the other hand, uncover hidden patterns and indirect relationships that may not be immediately apparent. By integrating both types of graphs, the model gains a richer and more nuanced understanding of the data, leading to improved feature representation and, consequently, better anomaly detection. The incorporation of LSTM networks plays a crucial role in capturing sequential dependencies within the data. Logs and other sequential data inherently possess temporal sequences that are critical for accurate anomaly detection. This model benefits from both temporal and spatial contexts, enhancing its ability to identify anomalies more accurately. Empirical results demonstrate the efficacy of this combined approach. The model consistently achieves higher Area Under the Curve (AUC) values, reflecting superior overall performance compared to other state-of-the-art methods. Moreover, the model attains higher True Positive Rates (TPR) while maintaining low False Positive Rates (FPR), illustrating its capability to detect anomalies with minimal false alarms accurately. Although CERT r5.2 and r6.2 serve as standard benchmarks for evaluating insider threat detection, their synthetic nature limits generalizability. Future work will focus on validating the proposed framework on additional real-world datasets to enhance its robustness and applicability.

\section{Data Sharing for Reproducibility}
We are making the code and related resources available in the public GitHub repository as follows. GitHub link: https://github.com/Yumlembam/Insider-Threat

\end{document}